# Predictive Maintenance using Machine Learning

A. Kane, A. Kore, A. Khandale, S. Nigade, P. P. Joshi

Computer Engineering Department, Pune Institute of Computer Technology

*Abstract*—Predictive maintenance (PdM) is a concept, which is implemented to effectively manage maintenance plans of the assets by predicting their failures with data driven techniques. In these scenarios, data is collected over a certain period of time to monitor the state of equipment. The objective is to find some correlations and patterns that can help predict and ultimately prevent failures. Equipment in manufacturing industry are often utilized without a planned maintenance approach. Such practise frequently results in unexpected downtime, owing to certain unexpected failures. In scheduled maintenance, the condition of the manufacturing equipment is checked after fixed time interval and if any fault occurs, the component is replaced to avoid unexpected equipment stoppages. On the flip side, this leads to increase in time for which machine is non-functioning and cost of carrying out the maintenance. The emergence of Industry 4.0 and smart systems have led to increasing emphasis on predictive maintenance (PdM) strategies that can reduce the cost of downtime and increase the availability (utilization rate) of manufacturing equipment. PdM also has the potential to bring about new sustainable practices in manufacturing by fully utilizing the useful lives of components.

## I. Introduction

Downtime of the equipment is the most serious issue in the manufacturing industries. Failure of the equipment results in downtime i.e. it remains useless until the maintenance is carried out and it starts functioning again. This scenario results in the decline of the productivity of the machine.

A common estimate is that almost every industry loses 5% to 20% of its productivity due to downtime and the maintenance costs can be outrageous. Machine failures can bring production to a screeching halt, and it costs an industry millions of dollars. Reducing downtime can help in improving overall efficiency of the industry.

ML techniques help us by assessing the historical data of the equipment over a fixed period of time and develop some patterns about degradation of the component. Thus we can predict the potential failure and schedule the maintenance before the system collapses entirely. This leads to minimum maintenance cost, productivity is maintained and proper utilization of the resources is ensured.

We propose to develop a Web application which takes machine sensor data as input to predict possibility of downtime of the machine as a result of the entire chain of neural constructions. This will be done by

1) Perform the data cleansing of the data set containing the records of different parameters collected by the sensors of the tubing machine and find the correlation to identify the pattern with respect to downtime of the tubing machine.
2) After analyzing the data set and deriving the correlation, identify and train the most suitable Machine Learning model which can predict the system failure well before in advance using the live data provided to the corresponding model.
3) The model will be very beneficial for the operator working on the machine and their respective managers and other stakeholders in terms of operations. The project will be very helpful in terms of saving time and provide highly efficient results. Hence the overall productivity and profit increases.

## II. Literature Survey

1) IJMOR 2020 Predictive Maintenance and Intelligent Sensors in Smart Factory: Review. Said Boutahari, Mohamed Ramadany, Soukaina Sadiki, Maurizio Facci, Driss Amegouz The paper reviews papers concerning predictive maintenance and intelligent sensors in smart factories. The paper's main contribution is the summary and overview of current trends in predictive maintenance in smart factories.
2) IEEE 2019 Predictive Maintenance 4.0 P. Po´or, J.Basl,. Zenisek Talks about the various levels of predictive maintenance. It also talks about the current stages certain industries are at in the implementation of predictive systems.
3) IEEE 2018 Forecasting faults of industrial equipment using machine learning classifiers N. Kolokas, T. Vafeiadis, D. Ioannidis, D. Tzovaras Predictive maintenance methodology to determine possible equipment stoppages along with the fault type in real time, utilizing sensor data from operation times are discussed. The warning time frame is set by the plant operations engineer to be as early as possible. Some widely used machine learning architectures are discussed.

## III. Overview

System failure is common occurrence across all the manufacturing industries. In most contexts, it is of utmost importance to predict the failure within time and take necessary steps.

Many manufacturing industries face the problem of the deterioration of their equipment, which becomes critical especially if this deterioration leads to a shutdown or unavailability of that component. Therefore a concept called Preventative Main-tenance (PM) emerged, which emphasizes on the prevention of asset failures and their surrounding environments.

Predictive maintenance using Machine Learning techniques tries to learn from data collected over a certain period of time and use live data to identify certain patterns of system failure, as opposed to conventional maintenance procedures relying on the life cycle of machine parts.

The ML-based predictive approach analyses the live data and tries to find out the correlation between certain parameters to predict the system failure or schedule maintenance of the equipment.

ML technology helps identify the fault lines by predicting the failures at the right time and thus utilizing resources effectively. This ensures the establishment of balance between maintenance needs and resource utilization.

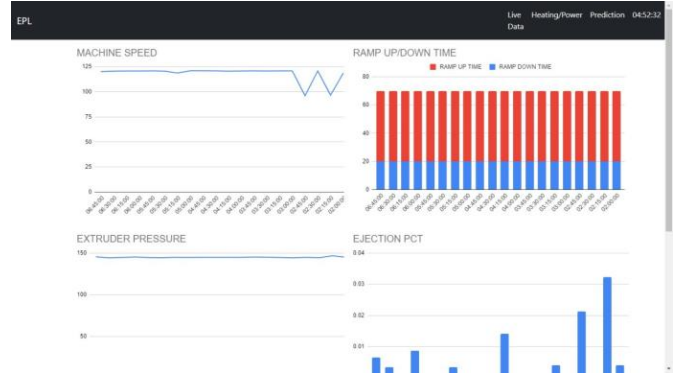

Fig. 2. Dashboard

## IV. MODULES

### A. Server

The data from various sensors of the tubing machine is fetched continuously and stored on the company server. Company uses MongoDB database since the volume of data generated is large. This data is then later used for the purpose of maintenance of the manufacturing unit. This project deals with small amount of data, hence uses MySQL database and uploads the data on the server.

### B. Client

The operator at a particular manufacturing plant can get the updates of the manufacturing unit. He would get to know the sensor readings at a particular timestamp from the server, which would help him to know if the unit is functioning properly or not, Hence, he can conduct the maintenance of the machine accordingly, thus preventing the complete machine failure.

### C. Backend

The data fetched from the sensors is cleaned and pre-processed to extract important features for data analysis and finding patterns and correlations among the parameters. The cleaned data is then used to train a machine learning model which would predict the parameter values over a period of time. Different models were trained on the parameters and their accuracy was calculated. The models were trained using classification and regression algorithms. LSTM model of deep learning was also implemented to make the predictions.

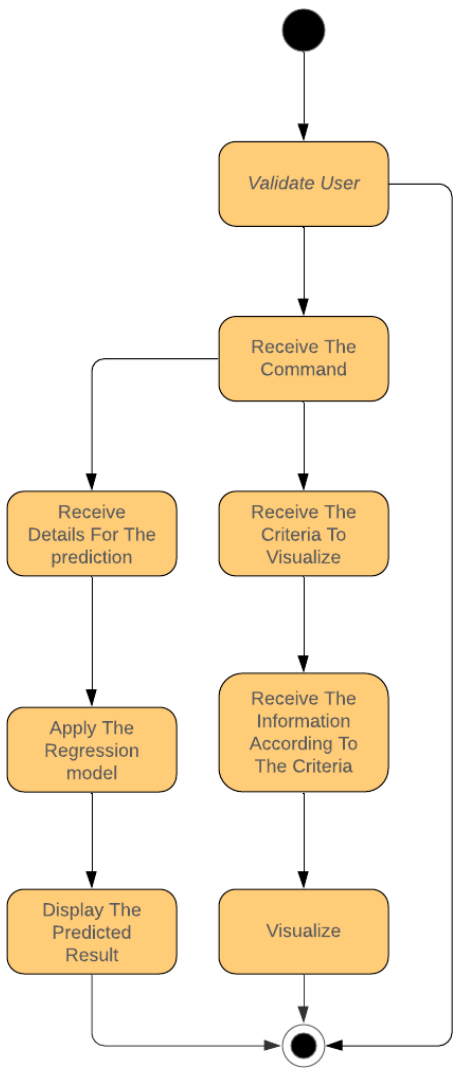

Fig. 1. Methodological Steps

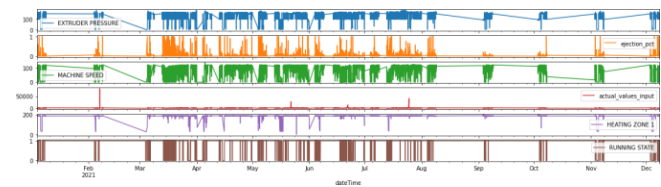

Fig. 3. Parameters

## V. IMPLEMENTATION DETAILS

### A. Dataset

The dataset used includes the sensor data of the machine MNL15 of which the main parameters considered are mentioned below:

1) EJECTION-PCT: Excess material ejected from machine during production.
2) EXTRUDER PRESSURE: Pressure sensor value from the extruder of the machine.
3) MACHINE SPEED: Speed of the machine while manufacturing.
4) ACTUAL-VALUES-INPUT: Input of raw material given to machine.
5) HEATING ZONE: Temperature reading of each individual section.

### B. LSTM

Long short-term memory is an artificial recurrent neural network (RNN) architecture used in the field of Natural Language Processing and Deep Learning. Standard feed forward neural networks such as CNN and RNN, don't have feedback connections. LSTMs have advantage in this aspect over simple neural networks. LSTMs can process single data points such as images as well as entire sequences of data such as speeches. The standard LSTM unit is made up of a cell, input gate, output gate and forget gate. The cell remembers the numbers periodically and three gates control the entry and exit of information from the cell. LSTMs have been established to address the vanishing gradient problem that can be encountered when training traditional RNNs. They can remember the information for long periods of time.

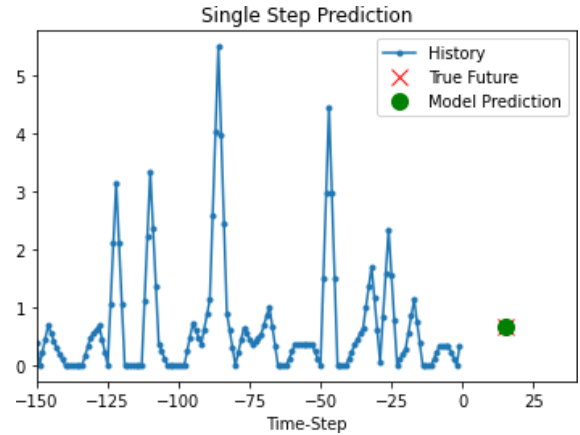

Fig. 5. Downtime Prediction by Model

revolutions per min, ams/ccm input etc. These parameters are recorded by the sensors of the tubing machine over a period of time, and hence maintenance can be scheduled as per the requirements. This model will prove to be extremely useful maintaining the productivity and minimizing the cost of maintenance. This will reduce the time for which the unit remains idle and maintenance. The model was very effective in predicting the time when the machine will be down based on the previous historical sensor data that it received.

## REFERENCES

[1] T. Salunkhe, N. I. Jamadar, and S. B. Kivade, "Prediction of Remaining Useful Life of Mechanical Components-A Review," vol. 3, no. 6, pp. 125–135, 2018.
[2] D. An, N. H. Kim, and J. H. Choi, "Practical options for selecting datadriven or physics-based prognostics algorithms with reviews," Reliab. Eng. Syst. Saf., vol. 133, pp. 223–236, 2019.
[3] N. Sapankevych, R. Sankar, "Time series prediction using support vector machines: A Survey", 2009.
[4] Angius, A.; Colledani, M.; Yemane, A. Impact of condition based maintenance policies on the service level of multi-stage manufacturing systems. Control. Eng. Pract. 2018, 76, 65–78.
[5] Ribeiro, I.M.; Godina, R.; Pimentel, C.; Silva, F.J.G.; Matias, J.C.O. Implementing TPM supported by 5S to improve the availability of an automotive production line. Procedia Manuf. 2019, 38, 1574–1581.
[6] Lasisi, A.; Attoh-Okine, N. Principal components analysis and track quality index: A machine learning approach. Transp. Res. Part. C Emerg. Technol. 2018, 91, 230–248.

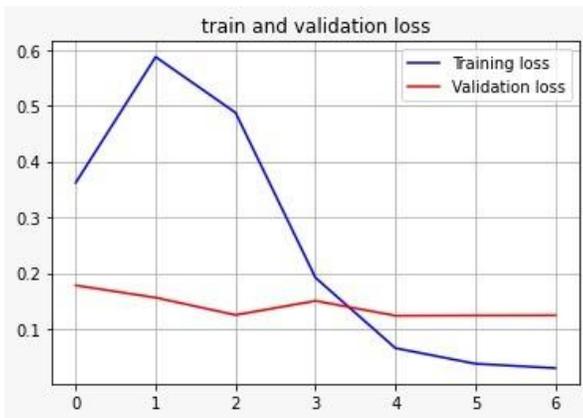

Fig. 4. Train vs Validation graph

## VI. CONCLUSION

This model proposes a system which will predict the failure of the manufacturing unit based on parameters like temperature, pressure and also the machine parameters like speed,